\documentclass[runningheads]{llncs}

\usepackage{eccv}

\usepackage{eccvabbrv}

\usepackage{multirow}
\usepackage{graphicx}
\usepackage{booktabs}
\usepackage{svg}

\usepackage[accsupp]{axessibility}  

\usepackage{hyperref}

\usepackage{orcidlink}

\usepackage{layouts}

\begin{document}

\title{Stipple: Real-Time Incremental Gaussian Splatting with Visual-Inertial Tracking}

\titlerunning{Stipple: Real-Time Incremental 3DGS with Visual-Inertial Tracking}

\author{Kilian Northoff\inst{1}\orcidlink{0009-0002-1876-9769} \and
  Mateo de Mayo\inst{1,2}\orcidlink{0009-0003-0729-3838} \and
  Daniel Cremers\inst{1,2}\orcidlink{0000-0002-3079-7984}}

\authorrunning{K.~Northoff et al.}

\institute{Technical University of Munich, Munich, Germany \and
  Munich Center for Machine Learning, Munich, Germany
}

\maketitle

\begin{abstract}
  3D Gaussian Splatting (3DGS) provides efficient rendering of photo-realistic scenes, but its heavy preprocessing and training steps make it a poor fit for applications that require real-time reconstruction in robotics or XR. This capability is important since it allows immediate feedback and interaction with new environments. Visual-inertial odometry (VIO) and simultaneous localization and mapping (VI-SLAM) systems, on the other hand, specifically target these real-time applications, which makes them a good choice for integration with 3DGS. We propose a new method that tracks and reconstructs simultaneously in real-time by leveraging an efficient visual-inertial tracking system based on Basalt together with a novel incremental method built on top of Brush, an efficient Rust-based GPU-vendor-agnostic implementation of 3D Gaussian Splatting. We show that many of the heavy preprocessing and training steps of 3DGS can be replaced with a more efficient incremental training strategy that has direct access to the information generated by the visual-inertial tracking system. Furthermore, we propose and combine multiple practical improvements to increase the efficiency of the training pipeline and adapt it to run in real-time,  parallel to the tracking thread. This work highlights the value of exploiting the complementary nature of SLAM and 3DGS, and how that can lead to promising results for real-time 3D reconstruction.
  \keywords{Gaussian Splatting \and VIO \and SLAM \and Robotics \and XR}
\end{abstract}

\section{Introduction}

Real-time reconstruction is crucial for applications in robotics, augmented
reality, and autonomous systems, where immediate feedback and interaction with
the environment are essential. While point-clouds are commonly used in real-time reconstruction, their sparse nature prevents them from recovering the fuller photometric information of the scene without expensive meshing and texturing steps. 3D Gaussian Splatting, however, provides a dense and photo-realistic representation of the scene, which can be used to improve the performance of these systems, especially in terms of perception and decision-making. By leveraging fully photometric (pixel-to-pixel) information, 3D Gaussian Splatting can provide a more accurate and detailed representation of the environment, which can be crucial for tasks such as object recognition, navigation, and manipulation.

Not only that, but for devices that explore new areas and are monitored by humans, e.g., rescue drones, exploration robots, or mixed reality headsets, this kind of representation can prove very useful for practitioners to understand the environment and make more informed decisions. Continuously updated digital twins are an important application on big construction sites and facilities, and, finally, a live 3D representation of the world is an outspoken goal of many frontier labs and companies. All of these applications require highly efficient and fast 3D reconstruction.

\begin{figure}[t]
  \centering
  \includegraphics[width=\textwidth]{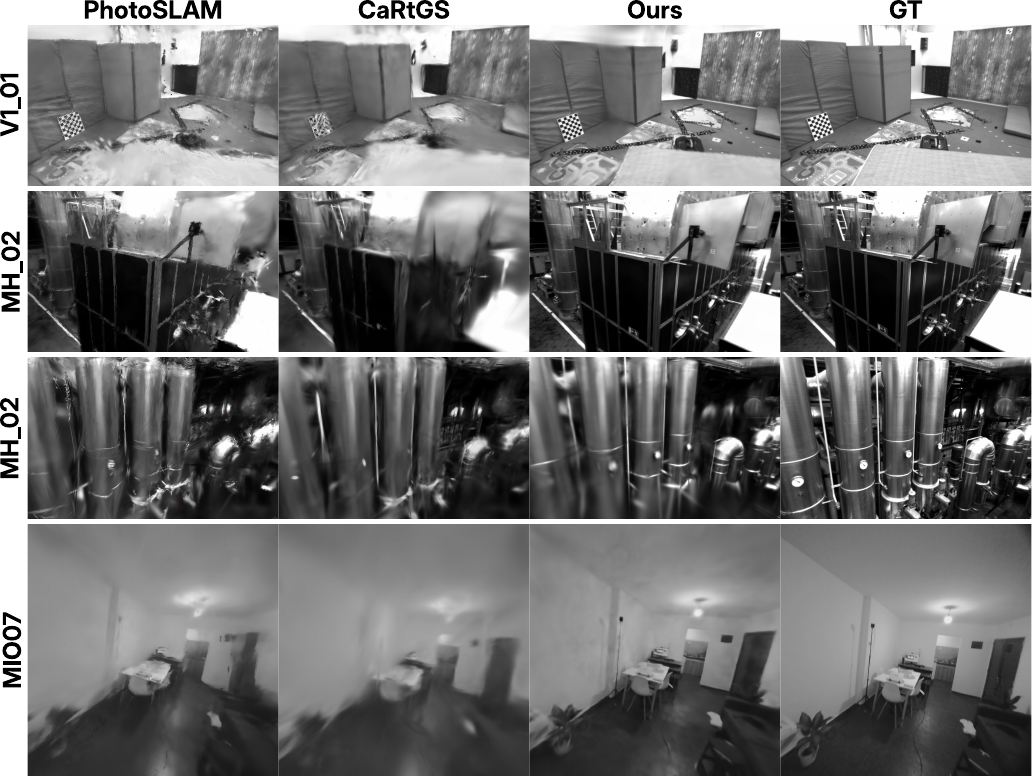}
  \caption{\textbf{Qualitative comparison against baselines.} We show a qualitative comparison of our method against the two most relevant baselines that support stereo input, PhotoSLAM~\cite{Huang2024photoslam} and CaRtGS~\cite{Feng2025cartgs}, on multiple sequences. All were run for the same amount of time (sequence length) on the same hardware. The results of our method provide better reconstruction quality for the same amount of time than the baselines. Notice that these are VI-SLAM datasets with monochrome tracking cameras, so the reconstructions are grayscale.}
  \label{fig:qualitative}
\end{figure}

However, while 3DGS boasts a very efficient rendering algorithm, the required preprocessing and training steps are too heavy for real-time usage. Preprocessing usually consists of running a heavy SfM pipeline like COLMAP on the input images to estimate camera poses and a point cloud of the scene for Gaussian initialization. Training is also very heavy, requiring many iterations of batched optimization to achieve a high-quality reconstruction.

We observe that many robotics, drone, and XR devices that require spatial understanding of their surroundings already use visual-inertial tracking and corresponding sensor setups. Given this, we propose a new method that leverages the efficiency of visual-inertial tracking systems and the rendering capabilities of 3D Gaussian Splatting to achieve real-time incremental reconstruction. We implement multiple improvements over the pipeline to speed up the training times including, but not limited to, fisheye support, stereo depth estimation, incremental training techniques, loop closure support, custom losses, and a more efficient Gaussian initialization method.

The contributions of this work are:
\begin{enumerate}
  \item A novel, GPU-agnostic, and real-time tracking and splatting pipeline that processes input stereo images and IMU data and generates camera poses and a 3DGS map of the environment simultaneously in real-time.
  \item Multiple VIO-aware techniques to make the 3D Gaussian Splatting pipeline efficient and suitable for real-time applications including, but not limited to, stereo depth estimation, keyframe culling, synchronization on pose graph optimization, and incremental training.
  \item Evaluation of the proposed method on relevant SLAM datasets compared against the two best relevant baselines that support stereo input, demonstrating its effectiveness and competitiveness in real-time reconstruction scenarios.
\end{enumerate}

The proposed method aims to bridge the gap between the high-quality rendering capabilities of 3D Gaussian Splatting and the real-time requirements of applications in robotics, augmented reality, and autonomous systems. By leveraging visual-inertial tracking and implementing efficient techniques, the method provides immediate reconstruction of new environments, which can enable new types of applications and interactions for spatial understanding. Furthermore, the reconstruction quality is superior to that of comparable methods as can be seen in Figure~\ref{fig:qualitative}.

\section{Related Work}

3D Gaussian Splatting (3DGS)~\cite{Kerbl2023gaussian} represents scenes as a collection of anisotropic Gaussian primitives rendered through a fast, differentiable tile-based rasterizer, achieving photo-realistic novel-view synthesis at real-time frame rates. Since rendering is fast but the underlying per-scene optimization is not, a separate line of work focuses purely on accelerating 3DGS training, for example by replacing its iterative densification with a multi-view-consistency-driven strategy~\cite{Ren2026fastgs} or with a one-step dense initialization from triangulated correspondences~\cite{Kotovenko2026edgs}. We build on Brush~\cite{Brussee2024brush}, a Rust and CubeCL implementation of 3DGS that avoids CUDA lock-in and runs cross-platform, and combine it with Basalt~\cite{Usenko2020visual}, a visual-inertial tracking system, with improvements from the Monado SLAM Dataset (MSD) benchmark~\cite{MSD}. It is similar to traditional visual(-inertial) SLAM systems such as ORB-SLAM3~\cite{Campos2021orbslam3} and DROID-SLAM~\cite{Teed2021droidslam}, but with a strong focus on resource efficiency and real-time performance.

\textbf{SLAM-based 3DGS.} A first wave of methods couples 3DGS directly with SLAM-style tracking and mapping, largely without a focus on real-time performance. SplaTAM~\cite{Keetha2024splatam} and MonoGS~\cite{Matsuki2024gaussian} were among the first to use Gaussians as the sole map representation for joint tracking and mapping, achieving strong visual quality but running at only a few frames per second. RTG-SLAM~\cite{Peng2024rtgslam} scales this idea to large scenes by forcing every Gaussian to be either an opaque, surface-fitting primitive or a transparent residual, getting close to real-time rates but still under 20\,fps. GS-SLAM~\cite{Yan2024gsslam} and CG-SLAM~\cite{Hu2024cgslam} add, respectively, an adaptive Gaussian expansion strategy and an uncertainty-aware Gaussian field derived from EWA splatting derivatives, while HF-SLAM~\cite{Sun2024hfslam} and MG-SLAM~\cite{Liu2025mgslam} target rendering-loss-guided densification and Manhattan-world-constrained hole filling, and DenseSplat~\cite{Li2025densesplat} uses a NeRF-based prior to densify maps built from sparse keyframes. HI-SLAM2~\cite{Zhang2025hislam2} instead targets the monocular RGB-only setting, combining easy-to-obtain monocular depth and normal priors with a learning-based tracking backend to reach reconstruction and rendering quality competitive with RGB-D methods. MM3DGS-SLAM~\cite{Sun2024mm3dgs} and VIGS SLAM~\cite{Pak2025vigsslam} additionally fuse IMU measurements into the tracking loss, but remain RGB-D-centric and are not real-time. Notably, and in contrast to our work, none of these methods uses stereo input, most require RGB-D depth (See Table I of \cite{Hu2025mgso}), and very few achieve performance comparable to real-time SLAM systems.

\textbf{Real-time SLAM-based 3DGS.} Closer to our goal, a smaller set of methods specifically targets real-time operation. Photo-SLAM~\cite{Huang2024photoslam} pairs ORB-SLAM3 tracking with a hybrid explicit/implicit Gaussian map and runs live on monocular, stereo, and RGB-D input, including on NVIDIA Jetson AGX Orin platforms (15-60W, 32-64GB). RGBD GS-ICP SLAM~\cite{Ha2024gsicp} shares a single Gaussian map between a Generalized-ICP tracker and the splatting-based mapper, reaching over 100\,fps for its tracking on a high end desktop (Ryzen 7 7800X3D, RTX 4090 24GB GPU), but is limited to RGB-D input. CaRtGS~\cite{Feng2025cartgs} improves only the splatting and densification side of the pipeline, and can be built on top of an existing tracker such as Photo-SLAM or RGBD GS-ICP SLAM. MGSO~\cite{Hu2025mgso} and GSO-SLAM~\cite{Yeon2026gsoslam} target monocular, photometric real-time SLAM by tightly coupling 3DGS with DSO~\cite{DSO}, MemGS~\cite{Bai2025memgs} focuses on reducing the memory footprint of the Gaussian map for NVIDIA Jetson platforms, and MonoGS++~\cite{Li2024monogspp} speeds up MonoGS by bootstrapping Gaussians from an external DPVO~\cite{DPVO} tracker. Of all these real-time methods, only Photo-SLAM and CaRtGS support stereo tracking cameras such as the ones found in the EuRoC~\cite{euroc}; most are monocular or RGB-D only, and are locked to high-power CUDA-capable GPUs. Our work, on the other hand, supports stereo-inertial input and multiple GPU backends.

We point the reader towards this survey work~\cite{Tosi2024neuslam} for a broader overview of the field and a more in-depth understanding of these gaps across the NeRF and 3DGS-based SLAM literature. To the best of our knowledge, our method is among the first to explicitly target real-time, near-instant reconstruction by combining cues from visual-inertial tracking with 3D Gaussian Splatting, to run on low-power devices without a CUDA-capable GPU, and to do so while achieving the highest tracking frame rates reported among comparable stereo-capable methods.

\section{Methodology}

\begin{figure}
  \centering
  \includegraphics[width=\textwidth]{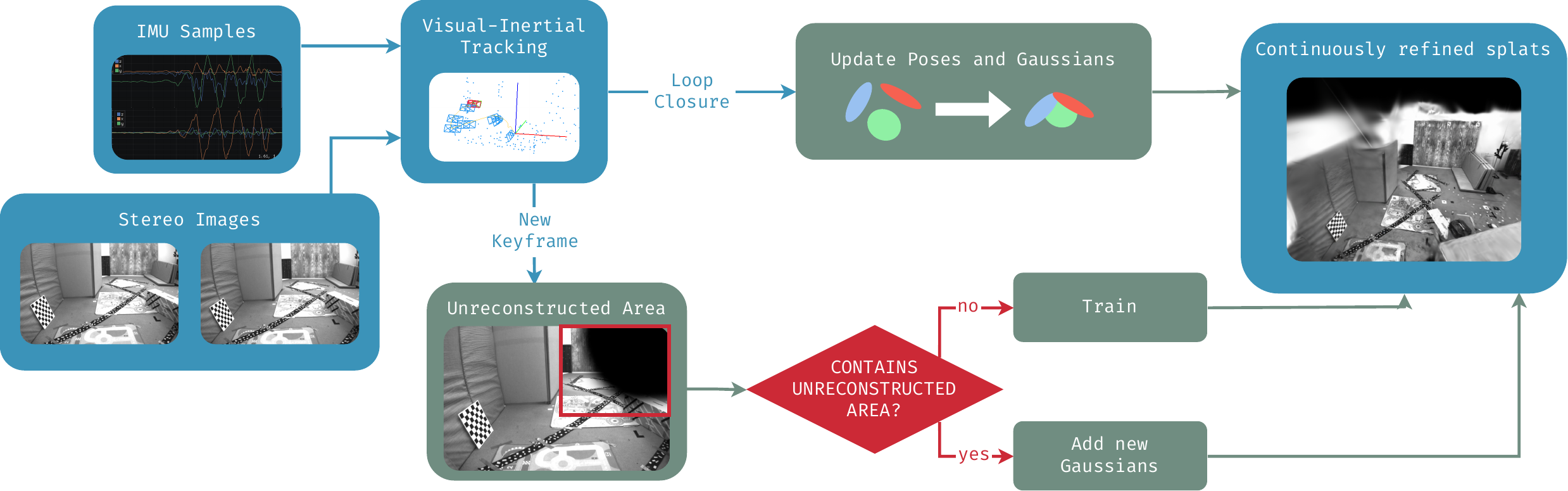}
  \caption{\textbf{System overview.} Keyframes and poses from the VI tracking system are submitted. Before committing a frame, the current map is rendered from that pose and scored for coverage and photometric error. If the view has unexplored or poorly reconstructed areas, it becomes an anchor frame, gets stereo depth, seeds new Gaussians, and corrects them on loop closure; otherwise it simply joins the ongoing optimization. All geometry is then refined by continuous optimization over every keyframe. Leveraging tracking poses, selected keyframes, loop closure signals, and stereo depth maps, the method achieves real-time incremental reconstruction.}
  \label{fig:method-overview}
\end{figure}

Basalt~\cite{Usenko2020visual,MSD} tracks every incoming stereo-inertial frame; images are undistorted and stereo-rectified with OpenCV before any further processing, which lets us support all OpenCV-supported camera models. Gaussians are only ever created from Basalt keyframes. Before a keyframe is committed, we render it from the current map and measure the fraction of pixels with alpha below $0.05$, i.e., the fraction of the view that is still unreconstructed. If this fraction exceeds $0.05$, the keyframe becomes an \emph{anchor frame}: it is assigned a stereo depth map and seeds new Gaussians (Section~\ref{sec:adding-gaussians}); otherwise it simply joins the pool of views used for ongoing optimization. Depth estimation (BM~\cite{BM}, SGBM-WLS~\cite{SGBMWLS,WLS}, or FoundationStereo~\cite{FoundationStereo}) is the pipeline's main computational bottleneck: FoundationStereo alone takes $\sim$150\,ms on an RTX 5060 Ti. Restricting it to anchor frames, typically only $\sim$10\% of all keyframes, is what keeps the pipeline real-time capable. We provide two figures relevant to the understanding of the pipeline: Figure~\ref{fig:method-overview} shows a high-level overview of the method, and Figure~\ref{fig:kfloop} shows the lifecycle of a single keyframe from the tracker to being part of the Gaussian reconstruction. There is no blocking between the tracking and the splatting threads: the SLAM thread never blocks on training, incoming views are queued for an asynchronous trainer, and a mutex serializes the only shared resource between the two, GPU access for stereo depth estimation and for splat optimization.

It is worth noting that we upstream\,\footnote{\url{https://github.com/ArthurBrussee/brush/pull/434}} support for the Kannala-Brandt~\cite{kb4} fisheye camera mode to Brush, with the objective of removing the need for undistortion and stereo rectification from the splatting, but we do not exercise this method in the experiments reported here and leave it for future work. Furthermore, we only consider stereo frames and leave out possibilities for exploiting multi-camera rigs fully like the ones encountered in the 4-camera \texttt{MG*} sequences of the Monado SLAM Dataset~\cite{MSD}. The side cameras are used in the tracking thread but currently ignored by the splatting; leveraging the higher field of view from these is a promising direction for future work.

\subsection{Adding Gaussians}
\label{sec:adding-gaussians}

\begin{figure}
  \centering
  \includegraphics[width=\textwidth]{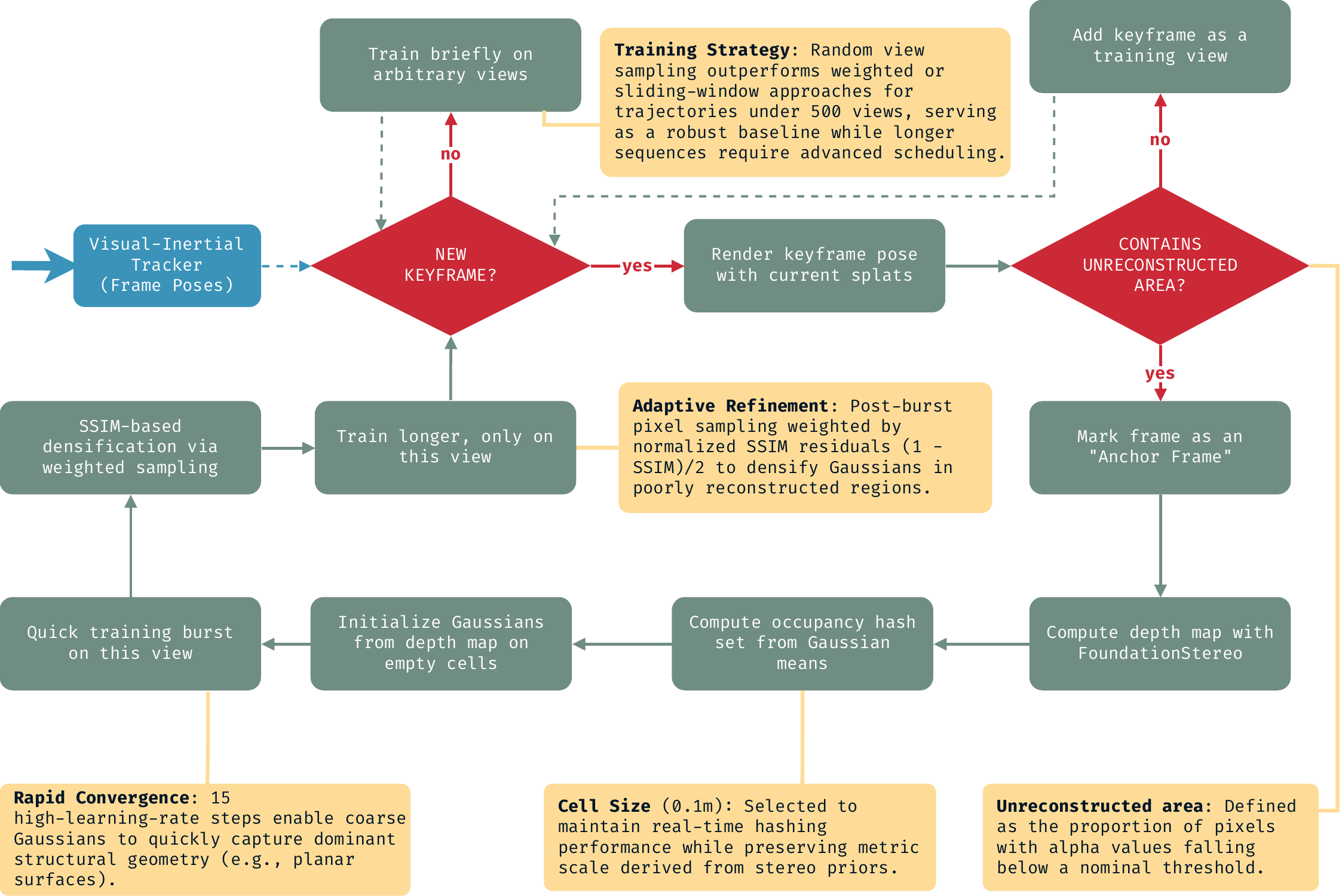}
  \caption{\textbf{Keyframe lifecycle.} New Gaussians are initialized from stereo-depth estimates, but placed only where space is not already occupied, so the map does not accumulate duplicates. A short training burst fits the new view; then photometric error steers extra Gaussians exactly where the render still disagrees with the image, before a final refinement.}
  \label{fig:kfloop}
\end{figure}

As shown in Figure~\ref{fig:kfloop}, once we have an anchor view selected, we proceed in the following way. \textbf{1. Depth-based seeding.} We compute a metric depth map with a stereo method: FoundationStereo~\cite{FoundationStereo} by default, and BM~\cite{BM} or SGBM-WLS~\cite{SGBMWLS,WLS} optionally. Then back-project each pixel into a 3D point using the known camera intrinsics and the anchor view's pose, yielding a dense set of candidate Gaussian means. \textbf{2. Occupancy pruning.} We then discard candidates that fall in already-reconstructed regions, using a CPU-side occupancy hashset built by hashing every existing Gaussian mean into a uniform grid (grid size $0.1$\,m in our experiments); a candidate survives only if its cell is not yet occupied. Coarser grids reduce hashset writes and thread synchronization at the cost of missing thinner structures, and this is what keeps the number of newly-seeded Gaussians small. \textbf{3. Burst fit.} We run a short optimization burst at an increased learning rate on only the new anchor frame, to see how well the sparse, freshly-seeded Gaussians already explain the image before allocating any more of them: flat, texture-poor regions (e.g., walls) are typically well reconstructed after this burst alone, while smaller and more complex structures are not. \textbf{4. SSIM-based densification.} After the burst, we compute the per-pixel SSIM between the rendered and observed anchor view (the same formulation used for evaluation) and derive the number of Gaussians to add from the mean SSIM,
$$
  n_\text{add} = \text{MAX\_SAMPLES} \cdot (1.0 - \overline{\text{SSIM}}) / 2,
$$
sampling that many pixel locations by weighted sampling with weights $w_i = 1.0 - \text{SSIM}_i$. We use SSIM rather than an L1/L2 photometric error to steer densification because it is bounded in $[-1, 1]$ and, more importantly, it captures local structural disagreement rather than the global consistency an L1 loss favors, which lets us target exactly the small, badly-reconstructed regions where the render still disagrees with the image. The new Gaussians are again unprojected using the depth map, with scales initialized from $k$-nearest-neighbor distances to existing Gaussians (as implemented in Brush), so that Gaussians in sparsely covered areas start out larger than those in already-dense regions. \textbf{5. Convergence.} We train for a further, longer interval on only the anchor view to let the newly added Gaussians converge.

Because a view only ever seeds Gaussians once, all Gaussians created from a given anchor frame remain contiguous in memory, which makes updating them on loop closure a simple, sequential operation (Section~\ref{sec:loop-closure}).

When no anchor frame is pending, we instead sample one keyframe uniformly at random among all keyframes seen so far and take a single training step on it. 
We ported the pose refinement from gsplat~\cite{Ye2024gsplat} into this optimization step, which is a local photometric alignment that does not feed back into the SLAM trajectory. We also experimented with a sliding window over recent keyframes and with frequency-weighted sampling that favors rarely-seen views, but an ablation showed uniform random sampling to perform best, a somewhat surprising result that we plan to investigate further.

\subsection{Loss Formulation}
\label{sec:loss-formulation}

The total loss is
$$
  \mathcal{L} = \lambda_1 \mathcal{L}_1 + \lambda_2\mathcal{L}_\text{SSIM} + \lambda_3\mathcal{L}_\text{depth} + \lambda_4\mathcal{L}_\text{anti-needle} + \lambda_5\mathcal{L}_\text{scale-regularizer},
$$
where $\mathcal{L}_1$ and $\mathcal{L}_\text{SSIM}$ follow the original 3DGS formulation~\cite{Kerbl2023gaussian}.

$\mathcal{L}_\text{depth}$, ported into Brush from gsplat~\cite{Ye2024gsplat}, is an L1 loss on inverse depth,
$$
  \mathcal{L}_\text{depth} = \left\lVert \frac{1}{D} - \frac{1}{\mathbb{E}\left[\hat{D}\right]} \right\rVert_1,
$$
with $D$ the depth map of the anchor view and $\hat{D}$ the rendered depth,
$$
  \hat{d_i} = \sum_{k \in G} d_k \alpha_k T_k, \qquad
  \mathbb{E}\left[\hat{d_i}\right] = \frac{\hat{d_i}}{\sum_{k \in G}\alpha_k},
$$
where $G$ is the set of Gaussians affecting pixel $i$, $d_k$ the depth of Gaussian $k$, $\alpha_k$ its alpha value, and $T_k$ its transmittance. Refer to 3DGS~\cite{Kerbl2023gaussian} for details on alpha blending. This term is only applied to anchor views, since a reference depth map $D$ only exists for anchor frames by construction: computing it for every keyframe would defeat the point of the anchor-frame distinction, which exists precisely to avoid running stereo depth estimation, the pipeline's main bottleneck, more often than necessary.

The anti-needle loss,
$$
  \mathcal{L}_\text{anti-needle} = \frac{1}{|G|} \sum_{i\in G} \frac{\max \mathbf{s}_i}{\min \mathbf{s}_i}
  = \frac{1}{|G|} \sum_{i\in G} \exp(\max \log\mathbf{s}_i - \min \log\mathbf{s}_i),
$$
with $G$ now the set of all Gaussians and $\mathbf{s}_i \in \mathbb{R}^3$ the scale of the covariance of Gaussian $i$ (stored as a logarithm, hence the right-hand side), regularizes Gaussians towards isotropic covariances, i.e., a less needle-like shape. This is a standard regularizer against highly anisotropic Gaussians, similar in spirit to the isotropic regularization used for example in MonoGS~\cite{Matsuki2024gaussian}: without it, Gaussians near the border of the reconstructed region tend to degenerate into needles that make the whole render look foggy.

Finally, our scale regularizer is defined as,
$$
  \mathcal{L}_\text{scale-regularizer} = \sum_{s \in S} s^2 \cdot \mathbb{I}\{s > \text{threshold}\},
$$
with $S$ the set of all Gaussian scales and $\mathbb{I}\{\cdot\}$ the indicator function. It is introduced to address issues originally observed on the EuRoC Machine Hall sequences where, otherwise, a subset of Gaussians explode in size and fog the entire scene.

\subsection{Loop Closure}
\label{sec:loop-closure}

Loop detection and pose-graph correction are handled entirely by the visual-inertial tracker and produce pose corrections for every keyframe that the splatting thread leverages. When such a correction arrives, every keyframe's stored pose is updated, since non-anchor keyframes are still used for training and their pose matters there. Only anchor frames additionally "own" Gaussians, so only their Gaussians receive a geometric update: we apply the same rigid transform that was applied to their anchor frame's pose, translating each Gaussian's mean and rotating its covariance accordingly.

\section{Experiments}

We evaluate on EuRoC~\cite{euroc} and on a selection of scenarios from the Monado SLAM Dataset~\cite{MSD}, excluding scenarios that contain gameplay to avoid dynamic occlusions; although the Reverb G2 headset used to record the latter has four cameras, they are only supported by our tracker component based on Basalt. However, our splatting thread and all the other baselines use only two front-facing sensors for both splatting and tracking since ORB-SLAM3 does not support setups with more than two cameras. For every method we report PSNR (dB), SSIM, LPIPS, tracking FPS, number of splats, and peak GPU memory, always on rectified images and with every sequence played back at its original recording rate so that no method is given extra wall-clock time to process a frame. We do not report tracking accuracy (ATE, RTE), since it is governed entirely by the underlying tracker and not by our contribution; we refer the reader to the ORB-SLAM3~\cite{Campos2021orbslam3}, Basalt~\cite{Usenko2020visual}, and Monado SLAM Dataset~\cite{MSD} papers for those numbers. All quantitative results report the best of 3 runs, with the same Basalt and Brush configuration used across every EuRoC and Monado scenario.

To ensure a fair comparison, we identified and fixed two issues in the public PhotoSLAM and CaRtGS implementations. First, both feed images to the tracking front-end slower than the sensor's original capture rate, which gives the splatting pipeline extra wall-clock time to process and train on each frame; since ORB-SLAM3 tracking is fast enough regardless, this does not affect tracking quality, but it does inflate reconstruction quality, so we corrected the feeding rate to match the original sequence timing for every method we compare against.

\subsection{Setup}

All experiments run on a single workstation with a Ryzen 9 9950X CPU and an NVIDIA GeForce RTX 5060 Ti (16\,GiB). We use Basalt~\cite{Usenko2020visual}, with the tracking improvements from~\cite{MSD} and a loop closure extension, as our stereo visual-inertial tracking system, implemented in C++, and Brush for Gaussian Splatting, implemented in Rust with GPU code in CubeCL. Since Brush does not depend on CUDA, and we provide BM and SGBM-WLS fallbacks for stereo depth alongside FoundationStereo, our pipeline can run on non-CUDA GPUs, unlike PhotoSLAM and CaRtGS.

\subsection{Qualitative Evaluation}

Figure~\ref{fig:qualitative} shows renders of our method against PhotoSLAM and CaRtGS after the same amount of wall-clock time on the same hardware. PhotoSLAM's reconstructions show a large number of needle-shaped, elongated Gaussians and ``flying'' groups of Gaussians, and distorted backgrounds and fine structure. CaRtGS's reconstructions are comparatively more coherent but still visibly distorted in several regions. Our method produces clearly sharper and more geometrically consistent reconstructions than either baseline for the same time budget.

\subsection{Quantitative Evaluation}

On the EuRoC Vicon Room sequences (Table~\ref{tab:euroc-vicon}) and on most Monado scenarios (Tables~\ref{tab:msd-mio}, \ref{tab:msd-mgo}, \ref{tab:msd-moo}), our method outperforms both PhotoSLAM and CaRtGS in PSNR, SSIM, and LPIPS, by margins of roughly $1$-$3$\,dB in PSNR, while tracking at $2$-$3\times$ their frame rate ($211$\,fps vs.\ $69$-$72$\,fps on Vicon Room; up to $264$\,fps on MSD). PhotoSLAM and CaRtGS also fail to produce usable results on any of the MOO sequences and on three of the four MGO sequences (Tables~\ref{tab:msd-mgo}, \ref{tab:msd-moo}), while our method completes all of them; on the one MGO sequence where a baseline result exists, our PSNR margin grows to $4.3$-$5.4$\,dB.

On the EuRoC Machine Hall sequences (Table~\ref{tab:euroc-mh}), however, the picture reverses: CaRtGS achieves the best PSNR and SSIM (21.80\,dB / 0.76 on average), while our method trails both baselines (19.48\,dB on average) despite still tracking at $173$\,fps, roughly $2.6\times$ CaRtGS's $64$\,fps. This coincides with a marked increase in splat count (394k on average, vs.\ 118k for CaRtGS) and peak GPU memory (4.7\,GB vs.\ 2.9\,GB), a pattern not present on the other scenarios. Our hypothesis is that the SSIM-based densification criterion (Section~\ref{sec:adding-gaussians}) over-triggers on Machine Hall, adding far more Gaussians than the fixed real-time training budget can converge, which lowers per-Gaussian quality; the scale-regularization loss (Section~\ref{sec:loss-formulation}) was introduced specifically to curb the resulting Gaussian-explosion pathology and partially compensates for it (Table~\ref{tab:ablation_study}), but does not fully close the gap. We leave a more principled fix, e.g.,\ a densification budget or an improved occupancy criterion, to future work.

While this evaluation protocol is in line with standard procedures, we note two limitations that we plan to address in future work: we evaluate on the same keyframes used to train each Gaussian splat, and this set of views differs between methods, since it depends on each method's own keyframe selection. A held-out, method-independent set of evaluation views would give a fairer picture and is a priority for a more complete version of this work.

\begin{table}
  \centering
  \caption{\textbf{EuRoC Vicon Room.} Best of 3 runs per method, equal running time. Stipple leads in most metrics while having comparable splat count and GPU memory.}
  \begin{tabular*}{\linewidth}{@{\extracolsep{\fill}}clcccccc|c}
    \toprule
    Method                          & Metric                         &     V1\_01     &     V1\_02     &     V1\_03     &     V2\_01     &     V2\_02     &     V2\_03     &      Avg       \\
    \midrule
    \multirow{6}{*}{\rotatebox[origin=c]{90}{PhotoSLAM}}   & PSNR $\uparrow$                &     22.02      &     22.60      &     17.90      &     20.62      &     20.87      &     17.33      &     20.22      \\
    & SSIM $\uparrow$                &      0.78      &      0.81      &      0.79      &      0.76      &      0.78      &      0.74      &      0.78      \\
    & LPIPS $\downarrow$             &      0.44      &      0.46      &      0.55      &      0.44      &      0.45      &      0.58      &      0.48      \\
    & Tracking FPS $\uparrow$        &       62       &       71       &       76       &       71       &       67       &       83       &       72       \\
    & \#Splats $\downarrow$          &      254k      &      224k      &      230k      &      261k      &      227k      &      172k      &      228k      \\
    & Peak GPU Mem (GB) $\downarrow$ &      2.5       &      2.1       &      2.9       &      2.8       &      2.7       &      3.4       &      2.8       \\
    \midrule
    \multirow{6}{*}{\rotatebox[origin=c]{90}{CaRtGS}}     & PSNR $\uparrow$                &     22.96      &     23.50      &     21.28      &     23.22      &     22.93      &     20.60      &     22.41      \\
    & SSIM $\uparrow$                &      0.80      &      0.83      &      0.84      &      0.83      &      0.83      &      0.80      &      0.82      \\
    & LPIPS $\downarrow$             &      0.44      &      0.44      &      0.50      &      0.39      &      0.41      &      0.52      &      0.45      \\
    & Tracking FPS $\uparrow$        &       57       &       68       &       72       &       68       &       66       &       81       &       69       \\
    & \#Splats $\downarrow$          &      79k       &  \textbf{60k}  &  \textbf{45k}  &      90k       &  \textbf{65k}  &  \textbf{50k}  &  \textbf{65k}  \\
    & Peak GPU Mem (GB) $\downarrow$ &  \textbf{2.3}  &  \textbf{1.9}  &  \textbf{2.8}  &      2.7       &  \textbf{2.3}  &      3.6       &  \textbf{2.6}  \\
    \midrule
    \multirow{6}{*}{\rotatebox[origin=c]{90}{\textbf{Ours}}} & PSNR $\uparrow$                & \textbf{25.10} & \textbf{23.78} & \textbf{22.00} & \textbf{24.48} & \textbf{23.30} & \textbf{22.28} & \textbf{23.49} \\
    & SSIM $\uparrow$                & \textbf{0.89}  & \textbf{0.85}  & \textbf{0.87}  & \textbf{0.85}  & \textbf{0.85}  & \textbf{0.85}  & \textbf{0.86}  \\
    & LPIPS $\downarrow$             & \textbf{0.32}  & \textbf{0.40}  & \textbf{0.42}  & \textbf{0.36}  & \textbf{0.38}  & \textbf{0.46}  & \textbf{0.39}  \\
    & Tracking FPS $\uparrow$        &  \textbf{208}  &  \textbf{203}  &  \textbf{221}  &  \textbf{203}  &  \textbf{206}  &  \textbf{223}  &  \textbf{211}  \\
    & \#Splats $\downarrow$          &  \textbf{52k}  &      99k       &      105k      &  \textbf{30k}  &      83k       &      124k      &      82k       \\
    & Peak GPU Mem (GB) $\downarrow$ &      2.8       &      3.2       &      3.1       &  \textbf{2.5}  &      3.0       &  \textbf{3.1}  &      2.9       \\
    \bottomrule
  \end{tabular*}
  \label{tab:euroc-vicon}
\end{table}

\begin{table}
  \centering
  \caption{\textbf{EuRoC Machine Hall.} Best of 3 runs per method, equal running time. Our method leads FPS by a wide margin but trails in PSNR/SSIM, with the largest splat count and GPU memory, from over-densification (Section~\ref{sec:loss-formulation}).}
  \label{tab:euroc-mh}
  \begin{tabular*}{\linewidth}{@{\extracolsep{\fill}}clccccc|c}
    \toprule
    Method                          & Metric                         &     MH\_01     &     MH\_02     &     MH\_03     &     MH\_04     &     MH\_05     &      Avg       \\
    \midrule
    \multirow{6}{*}{\rotatebox[origin=c]{90}{PhotoSLAM}}   & PSNR $\uparrow$                &     20.55      &     20.85      &     20.24      &     19.13      &     19.00      &     19.96      \\
    & SSIM $\uparrow$                &      0.69      &      0.69      &      0.69      &      0.71      &      0.70      &      0.70      \\
    & LPIPS $\downarrow$             &      0.42      &      0.43      &      0.47      &      0.45      &      0.45      &      0.44      \\
    & Tracking FPS $\uparrow$        &       61       &       65       &       64       &       69       &       70       &       66       \\
    & \#Splats $\downarrow$          &      379k      &      332k      &      339k      &      272k      &      314k      &      327k      \\
    & Peak GPU Mem (GB) $\downarrow$ &      3.2       &      2.9       &      2.8       &      3.1       &      3.3       &      3.1       \\
    \midrule
    \multirow{6}{*}{\rotatebox[origin=c]{90}{CaRtGS}}     & PSNR $\uparrow$                & \textbf{21.37} & \textbf{21.99} & \textbf{21.72} & \textbf{21.82} & \textbf{22.12} & \textbf{21.80} \\
    & SSIM $\uparrow$                & \textbf{0.73}  & \textbf{0.74}  & \textbf{0.74}  & \textbf{0.80}  & \textbf{0.79}  & \textbf{0.76}  \\
    & LPIPS $\downarrow$             & \textbf{0.40}  &      0.39      & \textbf{0.41}  & \textbf{0.37}  & \textbf{0.38}  & \textbf{0.39}  \\
    & Tracking FPS $\uparrow$        &       60       &       64       &       63       &       67       &       68       &       64       \\
    & \#Splats $\downarrow$          & \textbf{128k}  & \textbf{112k}  & \textbf{115k}  & \textbf{116k}  & \textbf{120k}  & \textbf{118k}  \\
    & Peak GPU Mem (GB) $\downarrow$ &  \textbf{3.0}  &  \textbf{2.7}  &  \textbf{2.6}  &  \textbf{3.1}  &  \textbf{3.2}  &  \textbf{2.9}  \\
    \midrule
    \multirow{6}{*}{\rotatebox[origin=c]{90}{\textbf{Ours}}} & PSNR $\uparrow$                &     19.37      &     21.39      &     18.18      &     18.93      &     19.54      &     19.48      \\
    & SSIM $\uparrow$                &      0.66      &      0.74      &      0.66      &      0.72      &      0.74      &      0.70      \\
    & LPIPS $\downarrow$             &      0.46      & \textbf{0.38}  &      0.47      &      0.44      &      0.43      &      0.44      \\
    & Tracking FPS $\uparrow$        &  \textbf{166}  &  \textbf{162}  &  \textbf{171}  &  \textbf{184}  &  \textbf{183}  &  \textbf{173}  \\
    & \#Splats $\downarrow$          &      234k      &      279k      &      611k      &      424k      &      420k      &      394k      \\
    & Peak GPU Mem (GB) $\downarrow$ &      3.7       &      4.1       &      5.4       &      5.1       &      4.9       &      4.7       \\
    \bottomrule
  \end{tabular*}
\end{table}

\begin{table}[t]
  \centering
  \caption{\textbf{MSD Valve Index.} Quantitative results on MIO sequences, best of 3 runs per method with the same running time. PhotoSLAM and CaRtGS failed to produce a result on MIO08. Our method attains the best PSNR, LPIPS, and tracking FPS on average, at roughly a third of the peak GPU memory of either baseline.}
  \begin{tabular*}{\linewidth}{@{\extracolsep{\fill}}clcccc|c}
    \toprule
    Method                        & Metric                         &     MIO05      &     MIO06      &     MIO07      &     MIO08      &      Avg       \\
    \midrule
    \multirow{6}{*}{\rotatebox[origin=c]{90}{PhotoSLAM}} & PSNR $\uparrow$                &     26.85      &     26.42      &     28.60      &       --       &     27.29      \\
    & SSIM $\uparrow$                &      0.94      &      0.94      &      0.95      &       --       &      0.95      \\
    & LPIPS $\downarrow$             &      0.39      &      0.39      &      0.36      &       --       &      0.38      \\
    & Tracking FPS $\uparrow$        &       48       &       50       &       55       &       --       &       51       \\
    & \#Splats $\downarrow$          &      63k       &      62k       &      47k       &       --       &      57k       \\
    & Peak GPU Mem (GB) $\downarrow$ &      9.0       &      7.4       &      6.4       &       --       &      7.6       \\
    \midrule
    \multirow{6}{*}{\rotatebox[origin=c]{90}{CaRtGS}}   & PSNR $\uparrow$                &     26.96      &     26.54      &     28.36      &       --       &     27.29      \\
    & SSIM $\uparrow$                &      0.94      &      0.95      &      0.95      &       --       &      0.95      \\
    & LPIPS $\downarrow$             &      0.40      &      0.39      &      0.38      &       --       &      0.39      \\
    & Tracking FPS $\uparrow$        &       43       &       48       &       46       &       --       &       46       \\
    & \#Splats $\downarrow$          &  \textbf{7k}   &  \textbf{6k}   &  \textbf{7k}   &       --       &  \textbf{6k}   \\
    & Peak GPU Mem (GB) $\downarrow$ &      7.9       &      6.5       &      6.6       &       --       &      7.0       \\
    \midrule
    \multirow{6}{*}{\rotatebox[origin=c]{90}{Ours}}    & PSNR $\uparrow$                & \textbf{28.28} & \textbf{28.31} & \textbf{29.27} & \textbf{27.32} & \textbf{28.30} \\
    & SSIM $\uparrow$                & \textbf{0.95}  & \textbf{0.95}  & \textbf{0.96}  & \textbf{0.95}  & \textbf{0.95}  \\
    & LPIPS $\downarrow$             & \textbf{0.35}  & \textbf{0.36}  & \textbf{0.34}  & \textbf{0.38}  & \textbf{0.36}  \\
    & Tracking FPS $\uparrow$        &  \textbf{143}  &  \textbf{146}  &  \textbf{145}  &  \textbf{150}  &  \textbf{146}  \\
    & \#Splats $\downarrow$          &      44k       &      51k       &      47k       &  \textbf{45k}  &      47k       \\
    & Peak GPU Mem (GB) $\downarrow$ &  \textbf{2.9}  &  \textbf{2.9}  &  \textbf{2.9}  &  \textbf{2.7}  &  \textbf{2.8}  \\
    \bottomrule
  \end{tabular*}
  \label{tab:msd-mio}
\end{table}

\begin{table}
  \centering
  \caption{\textbf{MSD HP Reverb G2.} Quantitative results on the MGO sequences, best of 3 runs per method with equal running time. PhotoSLAM and CaRtGS only yield a usable result on MGO05, where we lead by 4.3-5.4\,dB PSNR; ours is the only method to complete all four sequences. FPS speedup is smaller than on other MSD sequences because Basalt uses 4 cameras here for tracking, vs.\ 2 for the baselines.}
  \label{tab:msd-mgo}
  \begin{tabular*}{\linewidth}{@{\extracolsep{\fill}}clcccc|c}
    \toprule
    Method                        & Metric                         &     MGO05      &     MGO06      &     MGO07      &     MGO08      &      Avg       \\
    \midrule
    \multirow{6}{*}{\rotatebox[origin=c]{90}{PhotoSLAM}} & PSNR $\uparrow$                &     20.27      &       --       &       --       &       --       &     20.27      \\
    & SSIM $\uparrow$                &      0.75      &       --       &       --       &       --       &      0.75      \\
    & LPIPS $\downarrow$             &      0.62      &       --       &       --       &       --       &      0.62      \\
    & Tracking FPS $\uparrow$        &      103       &       --       &       --       &       --       &      103       \\
    & \#Splats $\downarrow$          &      61k       &       --       &       --       &       --       &      61k       \\
    & Peak GPU Mem (GB) $\downarrow$ &      3.4       &       --       &       --       &       --       &      3.4       \\
    \midrule
    \multirow{6}{*}{\rotatebox[origin=c]{90}{CaRtGS}}   & PSNR $\uparrow$                &     21.42      &       --       &       --       &       --       &     21.42      \\
    & SSIM $\uparrow$                &      0.78      &       --       &       --       &       --       &      0.78      \\
    & LPIPS $\downarrow$             &      0.63      &       --       &       --       &       --       &      0.63      \\
    & Tracking FPS $\uparrow$        &      102       &       --       &       --       &       --       &      102       \\
    & \#Splats $\downarrow$          &  \textbf{20k}  &       --       &       --       &       --       &  \textbf{20k}  \\
    & Peak GPU Mem (GB) $\downarrow$ &  \textbf{2.2}  &       --       &       --       &       --       &  \textbf{2.2}  \\
    \midrule
    \multirow{6}{*}{\rotatebox[origin=c]{90}{Ours}}    & PSNR $\uparrow$                & \textbf{25.70} & \textbf{22.69} & \textbf{25.67} & \textbf{25.60} & \textbf{24.91} \\
    & SSIM $\uparrow$                & \textbf{0.85}  & \textbf{0.81}  & \textbf{0.85}  & \textbf{0.83}  & \textbf{0.83}  \\
    & LPIPS $\downarrow$             & \textbf{0.51}  & \textbf{0.56}  & \textbf{0.51}  & \textbf{0.52}  & \textbf{0.52}  \\
    & Tracking FPS $\uparrow$        &  \textbf{156}  &  \textbf{152}  &  \textbf{157}  &  \textbf{160}  &  \textbf{156}  \\
    & \#Splats $\downarrow$          &      30k       &  \textbf{55k}  &  \textbf{41k}  &  \textbf{58k}  &      46k       \\
    & Peak GPU Mem (GB) $\downarrow$ &      2.4       &  \textbf{2.6}  &  \textbf{2.6}  &  \textbf{2.7}  &      2.6       \\
    \bottomrule
  \end{tabular*}
\end{table}

\begin{table}
  \centering
  \caption{\textbf{MSD Samsung Odyssey+.} Quantitative results on the Monado SLAM Dataset~\cite{MSD} MOO sequences, best of 3 runs with the same running time. Neither PhotoSLAM nor CaRtGS produced a usable result on any MOO sequence, so we report our method alone.}
  \begin{tabular*}{\linewidth}{@{\extracolsep{\fill}}clcccc|c}
    \toprule
    Method                        & Metric                         &     MOO05      &     MOO06      &     MOO07      &     MOO08      &      Avg       \\
    \midrule
    \multirow{6}{*}{\rotatebox[origin=c]{90}{Ours}}    & PSNR $\uparrow$                & {27.70} & {23.65} & {25.39} & {23.87} & {25.15} \\
    & SSIM $\uparrow$                & {0.89}  & {0.82}  & {0.86}  & {0.83}  & {0.85}  \\
    & LPIPS $\downarrow$             & {0.45}  & {0.54}  & {0.48}  & {0.52}  & {0.50}  \\
    & Tracking FPS $\uparrow$        &  {262}  &  {268}  &  {268}  &  {259}  &  {264}  \\
    & \#Splats $\downarrow$          &  {37k}  &  {42k}  &  {44k}  &  {55k}  &  {44k}  \\
    & Peak GPU Mem (GB) $\downarrow$ &  {2.5}  &  {2.6}  &  {2.5}  &  {2.7}  &  {2.6}  \\
    \bottomrule
  \end{tabular*}
  \label{tab:msd-moo}
\end{table}

\subsection{Ablation Study}

In Table~\ref{tab:ablation_study} we present a small ablation study on the EuRoC V1\_01 sequence, showing the effect of pose optimization (Section~\ref{sec:adding-gaussians}) and the max-splat-scale loss (Section~\ref{sec:loss-formulation}). Pose optimization alone accounts for most of the improvement (+0.84\,dB); the scale loss contributes a smaller, largely overlapping gain, and the two combined give the best result.

\begin{table}
  \centering
  \caption{\textbf{Ablation Study.} Ablation of pose optimization (Section~\ref{sec:adding-gaussians}) and the max-splat-scale loss (Section~\ref{sec:loss-formulation}), average PSNR over 3 runs on EuRoC V1\_01. Pose optimization alone accounts for most of the improvement (+0.84\,dB); the scale loss contributes a smaller, largely overlapping gain, and the two combined give the best result.}
  \label{tab:ablation_study}
  \begin{tabular}{c|c|c}
    \toprule
    Pose Optimization & Max Splat Scale Loss & PSNR           \\
    \midrule
                      &                      & 24.13          \\
                      & \checkmark           & 24.36          \\
    \checkmark        &                      & 24.97          \\
    \checkmark        & \checkmark           & \textbf{25.00} \\
    \bottomrule
  \end{tabular}
\end{table}

\section{Conclusion}

We presented a real-time, incremental 3D Gaussian Splatting pipeline that couples a stereo-inertial visual-inertial tracker with an efficient, GPU-vendor-agnostic 3DGS implementation. By reusing the poses, keyframes, and loop closures already produced by the tracking system, and by restricting the pipeline's main bottleneck, stereo depth estimation, to a small subset of anchor frames, our method reconstructs new environments in a matter of seconds while running concurrently with, and without blocking, the tracking thread. Across EuRoC and the Monado SLAM Dataset, our method tracks at $2$-$3\times$ the frame rate of the two strongest stereo-capable baselines, PhotoSLAM and CaRtGS, and matches or exceeds their reconstruction quality on most sequences, including several where the baselines fail to produce a usable result at all. The one clear exception is EuRoC Machine Hall, where a still-unexplained Gaussian-explosion pathology drives excessive densification and lowers quality relative to CaRtGS; we treat the scale-regularization loss introduced to curb it as a stop-gap rather than a solution. We hope this work encourages further exploration of the complementary nature of SLAM and 3D Gaussian Splatting, particularly for real-time robotics and XR applications where immediate, on-device reconstruction is necessary.

\section*{Acknowledgements}
This work was supported by the European Research Council (ERC) Advanced Grant
SIMULACRON, by the DFG project CR 250/26-1 “4D-YouTube”, by the GNI Project
“AI4Twinning”, and by the Munich Center for Machine Learning.

%
%
\bibliographystyle{splncs04}
\bibliography{main}

@String(CVPR  = {IEEE Conf. Comput. Vis. Pattern Recog.})

@String(ECCV  = {Eur. Conf. Comput. Vis.})

@String(TOG   = {ACM Trans. Graph.})

@String(CVPR  = {CVPR})

@String(ECCV  = {ECCV})

@String(TOG   = {ACM TOG})

@inproceedings{Bai2025memgs,
  title = {{{MemGS}}: {{Memory-Efficient Gaussian Splatting}} for {{Real-Time SLAM}}},
  shorttitle = {{{MemGS}}},
  booktitle = {2025 {{IEEE}}/{{RSJ International Conference}} on {{Intelligent Robots}} and {{Systems}} ({{IROS}})},
  author = {Bai, Yinlong and Zhang, Hongxin and Zhong, Sheng and Niu, Junkai and Li, Hai and He, Yijia and Zhou, Yi},
  year = 2025,
  month = oct,
  pages = {11097--11103},
  issn = {2153-0866},
  urldate = {2026-08-02}
}

@misc{Brussee2024brush,
  title = {Brush: {{A 3D Reconstruction Engine Using Gaussian Splatting}}},
  author = {Brussee, Arthur},
  year = 2026,
  urldate = {2025-08-26},
  copyright = {Apache-2.0}
}

@article{euroc,
  title = {The {{EuRoC}} Micro Aerial Vehicle Datasets},
  author = {Burri, M. and Nikolic, J. and Gohl, Pascal and Schneider, T. and Rehder, J. and Omari, Sammy and Achtelik, Markus and Siegwart, R.},
  year = 2016,
  journal = {Int. J. Robotics Res.}
}

@article{Campos2021orbslam3,
  title = {{{ORB-SLAM3}}: {{An Accurate Open-Source Library}} for {{Visual}}, {{Visual-Inertial}} and {{Multi-Map SLAM}}},
  shorttitle = {{{ORB-SLAM3}}},
  author = {Campos, Carlos and Elvira, Richard and Rodr{\'i}guez, Juan J. G{\'o}mez and Montiel, Jos{\'e} M. M. and Tard{\'o}s, Juan D.},
  year = 2021,
  month = dec,
  journal = {IEEE Transactions on Robotics},
  volume = {37},
  number = {6},
  eprint = {2007.11898},
  pages = {1874--1890},
  issn = {1552-3098, 1941-0468},
  urldate = {2021-12-13},
  archiveprefix = {arXiv}
}

@inproceedings{MSD,
  title = {The {{Monado SLAM Dataset}} for {{Egocentric Visual-Inertial Tracking}}},
  booktitle = {2025 {{IEEE}}/{{RSJ International Conference}} on {{Intelligent Robots}} and {{Systems}} ({{IROS}})},
  author = {{de Mayo}, Mateo and Cremers, Daniel and Pire, Taih{\'u}},
  year = 2025,
  month = oct,
  pages = {13111--13118},
  issn = {2153-0866},
  urldate = {2025-12-06}
}

@article{DSO,
  title = {Direct {{Sparse Odometry}}},
  author = {Engel, Jakob and Koltun, Vladlen and Cremers, Daniel},
  year = 2018,
  month = mar,
  journal = {IEEE Transactions on Pattern Analysis and Machine Intelligence},
  volume = {40},
  number = {3},
  pages = {611--625},
  issn = {0162-8828, 2160-9292},
  urldate = {2023-10-26},
  langid = {english}
}

@article{Feng2025cartgs,
  title = {{{CaRtGS}}: {{Computational Alignment}} for {{Real-Time Gaussian Splatting SLAM}}},
  shorttitle = {{{CaRtGS}}},
  author = {Feng, Dapeng and Chen, Zhiqiang and Yin, Yizhen and Zhong, Shipeng and Qi, Yuhua and Chen, Hongbo},
  year = 2025,
  month = may,
  journal = {IEEE Robotics and Automation Letters},
  volume = {10},
  number = {5},
  pages = {4340--4347},
  issn = {2377-3766},
  urldate = {2026-08-02}
}

@inproceedings{Ha2024gsicp,
  title = {{{RGBD GS-ICP SLAM}}},
  booktitle = {Computer {{Vision}} -- {{ECCV}} 2024},
  author = {Ha, Seongbo and Yeon, Jiung and Yu, Hyeonwoo},
  editor = {Leonardis, Ale{\v s} and Ricci, Elisa and Roth, Stefan and Russakovsky, Olga and Sattler, Torsten and Varol, G{\"u}l},
  year = 2025,
  pages = {180--197},
  publisher = {Springer Nature Switzerland},
  address = {Cham},
  isbn = {978-3-031-72764-1},
  langid = {english}
}

@article{SGBMWLS,
  title = {Stereo {{Processing}} by {{Semiglobal Matching}} and {{Mutual Information}}},
  author = {Hirschmuller, Heiko},
  year = 2008,
  month = feb,
  journal = {IEEE Transactions on Pattern Analysis and Machine Intelligence},
  volume = {30},
  number = {2},
  pages = {328--341},
  issn = {1939-3539},
  urldate = {2026-08-02}
}

@inproceedings{Huang2024photoslam,
  title = {Photo-{{SLAM}}: {{Real-Time Simultaneous Localization}} and {{Photorealistic Mapping}} for {{Monocular}}, {{Stereo}}, and {{RGB-D Cameras}}},
  shorttitle = {Photo-{{SLAM}}},
  booktitle = {2024 {{IEEE}}/{{CVF Conference}} on {{Computer Vision}} and {{Pattern Recognition}} ({{CVPR}})},
  author = {Huang, Huajian and Li, Longwei and Cheng, Hui and Yeung, Sai-Kit},
  year = 2024,
  month = jun,
  pages = {21584--21593},
  issn = {2575-7075},
  urldate = {2026-08-02}
}

@inproceedings{Hu2024cgslam,
  title = {{{CG-SLAM}}: {{Efficient Dense RGB-D SLAM}} in~a~{{Consistent Uncertainty-Aware 3D Gaussian Field}}},
  shorttitle = {{{CG-SLAM}}},
  booktitle = {Computer {{Vision}} -- {{ECCV}} 2024},
  author = {Hu, Jiarui and Chen, Xianhao and Feng, Boyin and Li, Guanglin and Yang, Liangjing and Bao, Hujun and Zhang, Guofeng and Cui, Zhaopeng},
  editor = {Leonardis, Ale{\v s} and Ricci, Elisa and Roth, Stefan and Russakovsky, Olga and Sattler, Torsten and Varol, G{\"u}l},
  year = 2025,
  pages = {93--112},
  publisher = {Springer Nature Switzerland},
  address = {Cham},
  isbn = {978-3-031-72698-9},
  langid = {english}
}

@inproceedings{Hu2025mgso,
  title = {{{MGSO}}: {{Monocular Real-Time Photometric SLAM}} with {{Efficient 3D Gaussian Splatting}}},
  shorttitle = {{{MGSO}}},
  booktitle = {2025 {{IEEE International Conference}} on {{Robotics}} and {{Automation}} ({{ICRA}})},
  author = {Hu, Yan Song and Abboud, Nicolas and Ali, Muhammad Qasim and Yang, Adam Srebrnjak and Elhajj, Imad and Asmar, Daniel and Chen, Yuhao and Zelek, John S.},
  year = 2025,
  month = may,
  pages = {11061--11067},
  urldate = {2026-08-02}
}

@article{kb4,
  title = {A Generic Camera Model and Calibration Method for Conventional, Wide-Angle, and Fish-Eye Lenses},
  author = {Kannala, J. and Brandt, S.S.},
  year = 2006,
  month = aug,
  journal = {IEEE Transactions on Pattern Analysis and Machine Intelligence},
  volume = {28},
  number = {8},
  pages = {1335--1340},
  issn = {1939-3539},
  urldate = {2026-08-02}
}

@inproceedings{Keetha2024splatam,
  title = {{{SplaTAM}}: {{Splat}}, {{Track}} \& {{Map 3D Gaussians}} for {{Dense RGB-D SLAM}}},
  shorttitle = {{{SplaTAM}}},
  booktitle = {2024 {{IEEE}}/{{CVF Conference}} on {{Computer Vision}} and {{Pattern Recognition}} ({{CVPR}})},
  author = {Keetha, Nikhil and Karhade, Jay and Jatavallabhula, Krishna Murthy and Yang, Gengshan and Scherer, Sebastian and Ramanan, Deva and Luiten, Jonathon},
  year = 2024,
  month = jun,
  pages = {21357--21366},
  issn = {2575-7075},
  urldate = {2026-08-02}
}

@article{Kerbl2023gaussian,
  title = {{{3D Gaussian Splatting}} for {{Real-Time Radiance Field Rendering}}},
  author = {Kerbl, Bernhard and Kopanas, Georgios and Leimkuehler, Thomas and Drettakis, George},
  year = 2023,
  month = jul,
  journal = {ACM Transactions on Graphics (TOG)},
  volume = {42},
  number = {4},
  pages = {139:1--139:14},
  issn = {0730-0301},
  urldate = {2026-08-02}
}

@inproceedings{BM,
  title = {Small {{Vision Systems}}: {{Hardware}} and {{Implementation}}},
  shorttitle = {Small {{Vision Systems}}},
  booktitle = {Robotics {{Research}}},
  author = {Konolige, Kurt},
  editor = {Shirai, Yoshiaki and Hirose, Shigeo},
  year = 1998,
  pages = {203--212},
  publisher = {Springer},
  address = {London},
  isbn = {978-1-4471-1580-9},
  langid = {english}
}

@misc{Kotovenko2026edgs,
  title = {{{EDGS}}: {{Eliminating Densification}} for {{Efficient Convergence}} of {{3DGS}}},
  shorttitle = {{{EDGS}}},
  author = {Kotovenko, Dmytro and Grebenkova, Olga and Ommer, Bj{\"o}rn},
  year = 2025,
  month = apr,
  number = {arXiv:2504.13204},
  eprint = {2504.13204},
  primaryclass = {cs},
  publisher = {arXiv},
  urldate = {2025-05-13},
  archiveprefix = {arXiv}
}

@article{Li2025densesplat,
  title = {{{DenseSplat}}: {{Densifying Gaussian Splatting SLAM With Neural Radiance Prior}}},
  shorttitle = {{{DenseSplat}}},
  author = {Li, Mingrui and Liu, Shuhong and Deng, Tianchen and Wang, Hongyu},
  year = 2026,
  month = feb,
  journal = {IEEE Transactions on Visualization and Computer Graphics},
  volume = {32},
  number = {2},
  pages = {1993--2006},
  issn = {1941-0506},
  urldate = {2026-08-02}
}

@misc{Li2024monogspp,
  title = {{{MonoGS}}++: {{Fast}} and {{Accurate Monocular RGB Gaussian SLAM}}},
  shorttitle = {{{MonoGS}}++},
  author = {Li, Renwu and Ke, Wenjing and Li, Dong and Tian, Lu and Barsoum, Emad},
  year = 2025,
  month = apr,
  number = {arXiv:2504.02437},
  eprint = {2504.02437},
  primaryclass = {cs.CV},
  publisher = {arXiv},
  urldate = {2026-08-02},
  archiveprefix = {arXiv}
}

@article{Liu2025mgslam,
  title = {{{MG-SLAM}}: {{Structure Gaussian Splatting SLAM With Manhattan World Hypothesis}}},
  shorttitle = {{{MG-SLAM}}},
  author = {Liu, Shuhong and Deng, Tianchen and Zhou, Heng and Li, Liuzhuozheng and Wang, Hongyu and Wang, Danwei and Li, Mingrui},
  year = 2025,
  journal = {IEEE Transactions on Automation Science and Engineering},
  volume = {22},
  pages = {17034--17049},
  issn = {1558-3783},
  urldate = {2026-08-02}
}

@inproceedings{Matsuki2024gaussian,
  title = {Gaussian {{Splatting SLAM}}},
  booktitle = {2024 {{IEEE}}/{{CVF Conference}} on {{Computer Vision}} and {{Pattern Recognition}} ({{CVPR}})},
  author = {Matsuki, Hidenobu and Murai, Riku and Kelly, Paul H. J. and Davison, Andrew J.},
  year = 2024,
  month = jun,
  pages = {18039--18048},
  issn = {2575-7075},
  urldate = {2026-08-02}
}

@article{WLS,
  title = {Fast {{Global Image Smoothing Based}} on {{Weighted Least Squares}}},
  author = {Min, Dongbo and Choi, Sunghwan and Lu, Jiangbo and Ham, Bumsub and Sohn, Kwanghoon and Do, Minh N.},
  year = 2014,
  month = dec,
  journal = {IEEE Transactions on Image Processing},
  volume = {23},
  number = {12},
  pages = {5638--5653},
  issn = {1941-0042},
  urldate = {2026-08-02}
}

@article{Pak2025vigsslam,
  title = {{{VIGS SLAM}}: {{IMU-Based Large-Scale RGB-D}} 3-{{D Gaussian Splatting SLAM}}},
  shorttitle = {{{VIGS SLAM}}},
  author = {Pak, Gyuhyeon and Kim, Euntai},
  year = 2026,
  journal = {IEEE Transactions on Instrumentation and Measurement},
  volume = {75},
  pages = {5010510--5010510},
  issn = {1557-9662},
  urldate = {2026-08-02}
}

@inproceedings{Peng2024rtgslam,
  title = {{{RTG-SLAM}}: {{Real-time 3D Reconstruction}} at {{Scale}} Using {{Gaussian Splatting}}},
  shorttitle = {{{RTG-SLAM}}},
  booktitle = {{{ACM SIGGRAPH}} 2024 {{Conference Papers}}},
  author = {Peng, Zhexi and Shao, Tianjia and Liu, Yong and Zhou, Jingke and Yang, Yin and Wang, Jingdong and Zhou, Kun},
  year = 2024,
  month = jul,
  series = {{{SIGGRAPH}} '24},
  pages = {1--11},
  publisher = {Association for Computing Machinery},
  address = {New York, NY, USA},
  urldate = {2026-08-01},
  isbn = {979-8-4007-0525-0}
}

@misc{Ren2026fastgs,
  title = {{{FastGS}}: {{Training 3D Gaussian Splatting}} in 100 {{Seconds}}},
  shorttitle = {{{FastGS}}},
  author = {Ren, Shiwei and Wen, Tianci and Fang, Yongchun and Lu, Biao},
  year = 2025,
  month = dec,
  number = {arXiv:2511.04283},
  eprint = {2511.04283},
  primaryclass = {cs.CV},
  publisher = {arXiv},
  urldate = {2026-08-02},
  archiveprefix = {arXiv}
}

@inproceedings{Sun2024hfslam,
  title = {High-{{Fidelity SLAM Using Gaussian Splatting}} with {{Rendering-Guided Densification}} and {{Regularized Optimization}}},
  booktitle = {2024 {{IEEE}}/{{RSJ International Conference}} on {{Intelligent Robots}} and {{Systems}} ({{IROS}})},
  author = {Sun, Shuo and Mielle, Malcolm and Lilienthal, Achim J. and Magnusson, Martin},
  year = 2024,
  month = oct,
  pages = {10476--10482},
  issn = {2153-0866},
  urldate = {2026-08-02}
}

@inproceedings{Sun2024mm3dgs,
  title = {{{MM3DGS SLAM}}: {{Multi-modal 3D Gaussian Splatting}} for {{SLAM Using Vision}}, {{Depth}}, and {{Inertial Measurements}}},
  shorttitle = {{{MM3DGS SLAM}}},
  booktitle = {2024 {{IEEE}}/{{RSJ International Conference}} on {{Intelligent Robots}} and {{Systems}} ({{IROS}})},
  author = {Sun, Lisong C. and Bhatt, Neel P. and Liu, Jonathan C. and Fan, Zhiwen and Wang, Zhangyang and Humphreys, Todd E. and Topcu, Ufuk},
  year = 2024,
  month = oct,
  pages = {10159--10166},
  issn = {2153-0866},
  urldate = {2026-08-02}
}

@inproceedings{DPVO,
  title = {Deep Patch Visual Odometry},
  booktitle = {Proceedings of the 37th {{International Conference}} on {{Neural Information Processing Systems}}},
  author = {Teed, Zachary and Lipson, Lahav and Deng, Jia},
  year = 2023,
  month = dec,
  series = {{{NIPS}} '23},
  pages = {39033--39051},
  publisher = {Curran Associates Inc.},
  address = {Red Hook, NY, USA},
  urldate = {2026-08-01}
}

@inproceedings{Teed2021droidslam,
  title = {{{DROID-SLAM}}: Deep Visual {{SLAM}} for Monocular, Stereo, and {{RGB-D}} Cameras},
  shorttitle = {{{DROID-SLAM}}},
  booktitle = {Proceedings of the 35th {{International Conference}} on {{Neural Information Processing Systems}}},
  author = {Teed, Zachary and Deng, Jia},
  year = 2021,
  month = dec,
  series = {{{NIPS}} '21},
  pages = {16558--16569},
  publisher = {Curran Associates Inc.},
  address = {Red Hook, NY, USA},
  urldate = {2026-08-01},
  isbn = {978-1-7138-4539-3}
}

@article{Tosi2024neuslam,
  title = {How {{NeRFs}} and 3-{{D Gaussian Splatting Are Reshaping SLAM}}: {{A Survey}}},
  shorttitle = {How {{NeRFs}} and 3-{{D Gaussian Splatting Are Reshaping SLAM}}},
  author = {Tosi, Fabio and Zhang, Youmin and Gong, Ziren and Sandstr{\"o}m, Erik and Mattoccia, Stefano and Oswald, Martin R. and Poggi, Matteo},
  year = 2026,
  journal = {IEEE Transactions on Robotics},
  volume = {42},
  pages = {1405--1427},
  issn = {1941-0468},
  urldate = {2026-08-02}
}

@article{Usenko2020visual,
  title = {Visual-{{Inertial Mapping}} with {{Non-Linear Factor Recovery}}},
  author = {Usenko, Vladyslav and Demmel, Nikolaus and Schubert, David and St{\"u}ckler, J{\"o}rg and Cremers, Daniel},
  year = 2020,
  month = apr,
  journal = {IEEE Robotics and Automation Letters},
  volume = {5},
  number = {2},
  eprint = {1904.06504},
  pages = {422--429},
  issn = {2377-3766, 2377-3774},
  urldate = {2021-11-05},
  archiveprefix = {arXiv},
  langid = {english}
}

@inproceedings{FoundationStereo,
  title = {{{FoundationStereo}}: {{Zero-Shot Stereo Matching}}},
  shorttitle = {{{FoundationStereo}}},
  booktitle = {2025 {{IEEE}}/{{CVF Conference}} on {{Computer Vision}} and {{Pattern Recognition}} ({{CVPR}})},
  author = {Wen, Bowen and Trepte, Matthew and Aribido, Joseph and Kautz, Jan and Gallo, Orazio and Birchfield, Stan},
  year = 2025,
  month = jun,
  pages = {5249--5260},
  issn = {2575-7075},
  urldate = {2026-08-02}
}

@inproceedings{Yan2024gsslam,
  title = {{{GS-SLAM}}: {{Dense Visual SLAM}} with {{3D Gaussian Splatting}}},
  shorttitle = {{{GS-SLAM}}},
  booktitle = {2024 {{IEEE}}/{{CVF Conference}} on {{Computer Vision}} and {{Pattern Recognition}} ({{CVPR}})},
  author = {Yan, Chi and Qu, Delin and Xu, Dan and Zhao, Bin and Wang, Zhigang and Wang, Dong and Li, Xuelong},
  year = 2024,
  month = jun,
  pages = {19595--19604},
  issn = {2575-7075},
  urldate = {2026-08-02}
}

@article{Ye2024gsplat,
  title = {Gsplat: {{An}} Open-Source Library for {{Gaussian}} Splatting},
  author = {Ye, Vickie and Li, Ruilong and Kerr, Justin and Turkulainen, Matias and Yi, Brent and Pan, Zhuoyang and Seiskari, Otto and Ye, Jianbo and Hu, Jeffrey and Tancik, Matthew and Kanazawa, Angjoo},
  year = 2024,
  journal = {arXiv preprint arXiv:2409.06765},
  eprint = {2409.06765},
  primaryclass = {cs.CV},
  archiveprefix = {arXiv}
}

@article{Yeon2026gsoslam,
  title = {{{GSO-SLAM}}: {{Bidirectionally Coupled Gaussian Splatting}} and {{Direct Visual Odometry}}},
  shorttitle = {{{GSO-SLAM}}},
  author = {Yeon, Jiung and Ha, Seongbo and Yu, Hyeonwoo},
  year = 2026,
  month = apr,
  journal = {IEEE Robotics and Automation Letters},
  volume = {11},
  number = {4},
  pages = {5033--5040},
  issn = {2377-3766},
  urldate = {2026-08-02}
}

@article{Zhang2025hislam2,
  title = {{{HI-SLAM2}}: {{Geometry-Aware Gaussian SLAM}} for {{Fast Monocular Scene Reconstruction}}},
  shorttitle = {{{HI-SLAM2}}},
  author = {Zhang, Wei and Cheng, Qing and Skuddis, David and Zeller, Niclas and Cremers, Daniel and Haala, Norbert},
  year = 2025,
  journal = {IEEE Transactions on Robotics},
  volume = {41},
  pages = {6478--6493},
  issn = {1941-0468},
  urldate = {2026-08-02}
}
\end{document}